\documentclass[10pt,journal,cspaper,compsoc]{IEEEtran}

\usepackage{epsfig}
\usepackage{graphicx}
\usepackage{amsmath}
\usepackage{amssymb}
\usepackage{latexsym, bm}
\usepackage{algorithm}
\usepackage{color}
\usepackage{url}

\newcommand{\etal}{\textit{et al}.}
\newcommand{\ie}{\textit{i}.\textit{e}.}

\newcommand{\Eref}[1]{Equation~(\ref{#1})}
\newcommand{\Fref}[1]{Figure~\ref{#1}}

\newcommand{\eref}[1]{Eq.~(\ref{#1})}
\newcommand{\fref}[1]{Fig.~\ref{#1}}

\newcommand{\argmin}{\operatornamewithlimits{argmin}}

\ifCLASSOPTIONcompsoc
\else
\fi

\ifCLASSINFOpdf
\else
\fi

\begin{document}

\title{Multiview Rectification of Folded Documents}

\author{Shaodi~You,~\IEEEmembership{Member,~IEEE,}
        Yasuyuki~Matsushita,~\IEEEmembership{Member,~IEEE,}
        Sudipta~Sinha,~\IEEEmembership{Member,~IEEE}
		Yusuke~Bou,~\IEEEmembership{Member,~IEEE,}
		Katsushi~Ikeuchi,~\IEEEmembership{Fellow,~IEEE,}
\IEEEcompsocitemizethanks{
\IEEEcompsocthanksitem S. You is with National ICT Australia (NICTA), and Australian National University, Australia.\protect~E-mail: youshaodi@gmail.com
\IEEEcompsocthanksitem Y. Matsushita is with Osaka University, Japan.\protect
\\E-mail: yasumat@ist.osaka-u.ac.jp
\IEEEcompsocthanksitem S. Sinha is with Microsoft Research Redmond, USA.\protect
\\E-mail: sudipta.sinha@microsoft.com
\IEEEcompsocthanksitem Y. Bou is with Microsoft, Japan.\protect
~E-mail: yusuketa@microsoft.com
\IEEEcompsocthanksitem K. Ikeuchi is with Microsoft Research Asia, China.\protect
\\E-mail: ki@cvl.iis.u-tokyo.ac.jp
}
}

\markboth{IEEE TRANSACTION ON PATTERN ANALYSIS AND MACHINE INTELLIGENCE}%
{Shell \MakeLowercase{\textit{et al.}}: Bare Demo of IEEEtran.cls for Computer Society Journals}

\IEEEcompsoctitleabstractindextext{%
\begin{abstract}
Digitally unwrapping images of paper sheets is crucial for accurate document scanning and text recognition. This paper presents a method for automatically rectifying curved or folded paper sheets from a few images captured from multiple viewpoints.
Prior methods either need expensive 3D scanners or model deformable surfaces
using over-simplified parametric representations. In contrast, our method uses regular images and is based on general developable surface models that can represent a wide variety of paper deformations. Our main contribution is a new robust rectification method based on ridge-aware 3D reconstruction of a paper sheet and unwrapping the reconstructed surface using properties of developable surfaces via $\ell_1$ conformal mapping. We present results on several examples including book pages, folded letters and shopping receipts.
\end{abstract}

}

\maketitle

\IEEEdisplaynotcompsoctitleabstractindextext

 \ifCLASSOPTIONpeerreview
 \begin{center} \bfseries EDICS Category: 3-BBND \end{center}
 \fi

\IEEEpeerreviewmaketitle

\section{Introduction}

Digitally scanning paper documents for sharing and editing is becoming a common daily task.
Such paper sheets are often curved or folded, and proper rectification is important for high-fidelity digitization and text recognition.
Flatbed scanners allow physical rectification of such documents but are not suitable for hardcover books. 
For a wider applicability of document scanning, it is wanted a flexible technique for digitally rectifying folded documents.

There are two major challenges in document image rectification.
First, for a proper rectification, the 3D shape of curved and folded paper sheets must be estimated. Second, the estimated surface must be flattened without introducing distortions.
Prior methods for 3D reconstruction of curved paper sheets either use specialized hardware~\cite{Pilu01,Meng14,Brown04} or assume simplified parametric shapes~\cite{Wada97,Zhang09, Tsoi07, Koo09, Stamatopoulos11, Zhang11, Meng14}, such as generalized cylinders (\fref{Fig:Ruler}a). However, these methods are difficult to use due to bulky hardware or make restrictive assumptions about the deformations of the paper sheet.


\begin{figure*}[tbhp]
	\centering
	\includegraphics[width=\linewidth]{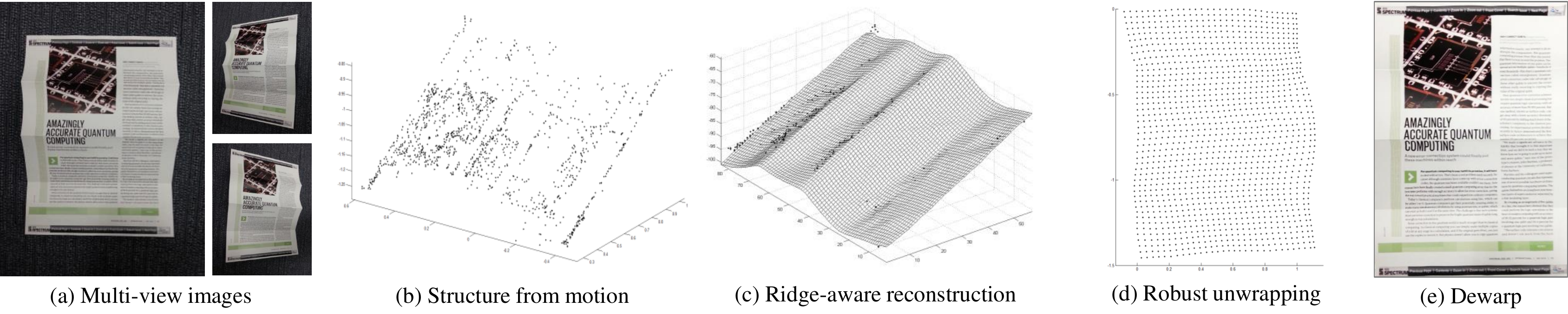}
	\vspace{-18pt}
	\caption{
	\small{
		Our technique recovers a ridge-aware 3D reconstruction of the document surface from a sparse 3D point cloud. The final rectified image is then obtained via robust conformal mapping.
	}
	}
	\label{Fig:Pipeline}
	\vspace{-12pt}
\end{figure*}

In this paper, we present a convenient method for digitally rectifying heavily curved and folded paper sheets from a few uncalibrated images captured with a hand-held camera from multiple viewpoint.
Our method uses structure from motion (SfM) to recover an initial sparse 3D point cloud from the uncalibrated images.
To accurately recover the dense 3D shape of paper sheet without losing high-frequency structures such as folds and creases, we develop a \emph{ridge-aware} surface reconstruction method. Furthermore, to achieve robustness to outliers present in the sparse SfM 3D point cloud caused by repetitive document textures, we pose the surface reconstruction task as a robust Poisson surface reconstruction based on $\ell_1$ optimization.
Next, to unwrap the reconstructed surface, we propose a robust conformal mapping method by incorporating ridge-awareness priors and $\ell_1$ optimization technique. See~\fref{Fig:Pipeline} for an overview.

The contributions of our work are threefold.
First, we show how ridge-aware regularization can be used for both 3D surface reconstruction and flattening (conformal mapping) to improve accuracy. Our ridge-aware reconstruction method preserves the sharp structure of folds and creases. Ridge-awareness priors act as non-local regularizers that reduce global distortions during the surface flattening step.
Second, we extend the Poisson surface reconstruction~\cite{Kazhdan06} and least-squares conformal mapping (LSCM)~\cite{Levy02} algorithms by explicitly dealing with outliers using $\ell_1$ optimization. Finally, we describe a practical system for rectifying curved and folded documents that can be used with ordinary digital cameras.

\section{Related Work}

\begin{figure}[t] 
	\centering
	\includegraphics[width=\linewidth]{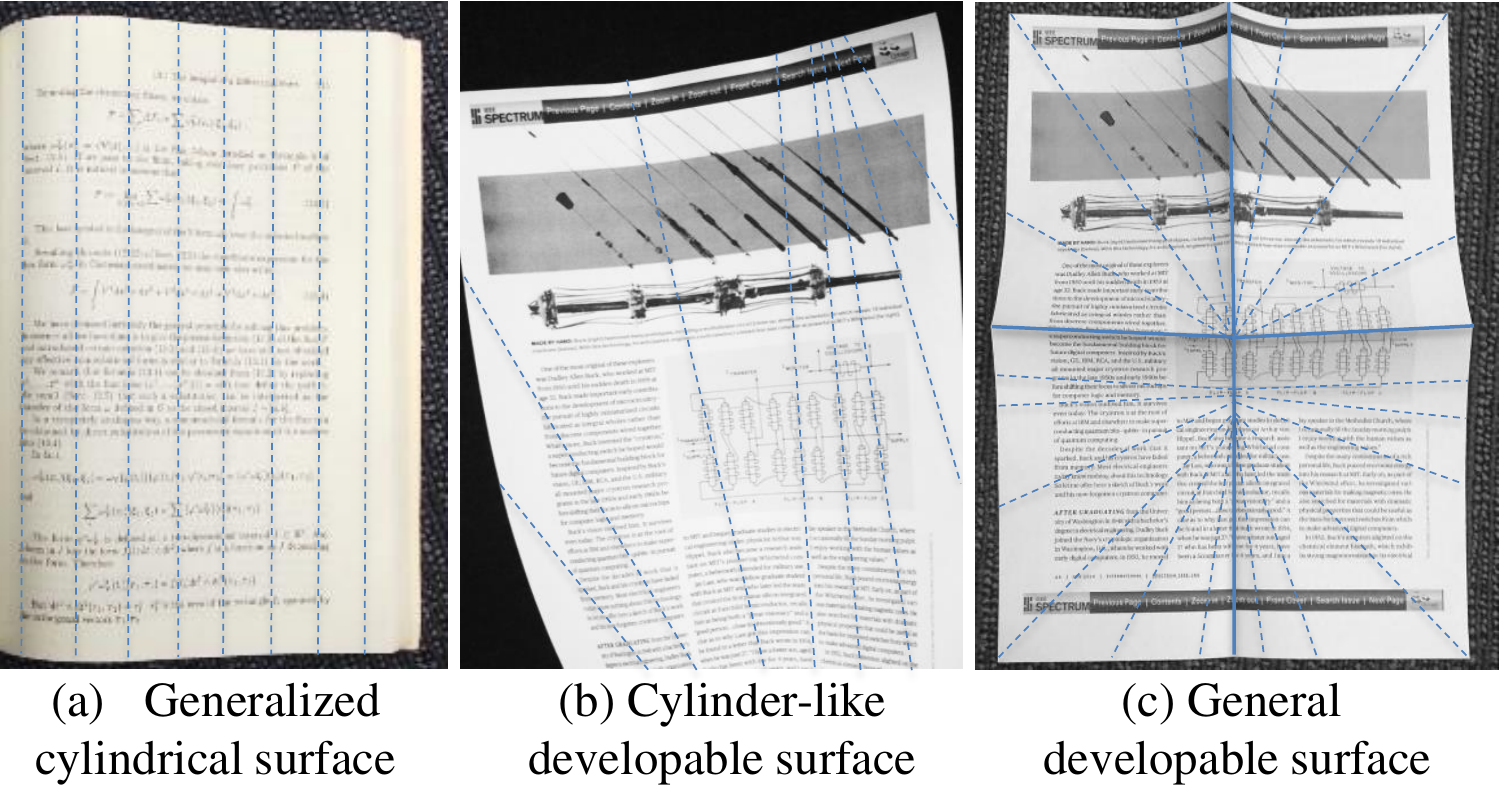}
	\vspace{-18pt}
	\caption{
		\small{
		Developable surfaces with underlying rulers (lines with zero Gaussian curvature) and fold lines (ridges) shown as dotted and solids lines respectively.
		Examples of (a) smooth parallel rulers, (b) smooth rulers not parallel to each other and (c) rulers and ridges in arbitrary directions.
	}
}
	\label{Fig:Ruler}
	\vspace{-18pt}
\end{figure}

The topic of digital rectification of curved and folded documents has been actively studied in both the computer vision and document processing communities.
It is common to model paper sheets as developable surfaces which have underlying rulers corresponding to lines with zero Gaussian curvature. Many existing methods assume generalized cylindrical surfaces where the paper is curved only in one direction and thus can be parameterized using a 1D smooth function. Such surfaces do not require an explicit parameterization of the rulers. See \fref{Fig:Ruler}a for an example. A variety of existing techniques recover surface geometry using this assumption.
Shape from shading methods were first used by Wada~\etal~\cite{Wada97}, Tan~\etal ~\cite{Zhang04, Tan06}, Courteille~\etal~\cite{Courteille07} and Zhang~\etal~\cite{Zhang09} whereas
shape from boundary methods were explored by Tsoi~\etal~\cite{Tsoi04, Tsoi07}.
Binocular stereo matching with calibrated cameras was used by Yamashita~\etal~\cite{Yamashita04}, Koo~\etal~\cite{Koo09} and Tsoi~\etal ~\cite{Tsoi07}.
Shape from text lines is another popular method for reconstructing the document surface geometry~\cite{Cao03, Zhang04, Ezaki05, Ulges05, Lu06, Fu07, Stamatopoulos11, Meng12, Zhang11, Liu14, Kim15, Salvi15}. However, these methods assume that the document contains well-formatted printed characters.

Some recent methods relax the parallel ruler assumption (see \fref{Fig:Ruler}b).
However, the numerous parameters in these models makes the optimization quite challenging.
Liang~\etal~\cite{Liang08} and Tian~\etal~\cite{Tian11} use text lines. Although these methods can handle a single input image, the strong assumptions on surface geometry, contents and illumination limit the applicability.
Meng~\etal designed a special calibrated active structural light device to retrieve the two parallel 1D curvatures~\cite{Meng14}, the surface can be parameterized by assuming appropriate boundary conditions and constraints on ruler orientations.
Perriollat \etal\cite{Perriollat13} use sparse SfM points but assume they are reasonably dense and well distributed. Their parameterization is sensitive to noise and can be unreliable when the 3d point cloud is sparse or has varying density.

For rectification of documents with arbitrary distortion and content (\fref{Fig:Ruler}c), other methods require specialized devices and use non-parametric approaches.
Brown~\etal~\cite{Brown04} use a calibrated mirror system to obtain 3D geometry using multi-view stereo. They unwrap the reconstructed surface using constraints on elastic energy, gravity and collision.  The model is not ideal for paper documents because developable surfaces are not elastic. Later, they propose using dense 3D range data~\cite{Brown07} after which they flatten the surface using least square conformal mapping~\cite{Levy02}.
Zhang~\etal~\cite{Zhang08} also use dense range scans and use rigid constraints instead of elastic constraints with the method proposed in~\cite{Brown04}.
Pilu~\cite{Pilu01} assumes that a dense 3D mesh is available and minimizes the global bending potential energy to flatten the surface.
None of these existing methods are as practical and convenient as our method that only requires a hand-held camera and a few images.

\vspace{-6pt}
\section{Proposed Method}
\vspace{-3pt}

Our method has two main steps -- 3D document surface reconstruction and unwrapping of the reconstructed surface.
For now, we assume that a set of sparse 3D points on the surface are available.
Next, we describe our new algorithms for ridge-aware surface reconstruction and robust surface unwrapping.

\vspace{-6pt}
\subsection{Ridge-aware surface reconstruction}
\vspace{-3pt}

Dense methods are favored for 3D scanning of folded and curved documents~\cite{Avron10, Oztireli09}.
This is because existing methods for surface reconstruction from sparse 3D points tend to produce excessive
smoothing and fail to preserve sharp creases and folds ie. ridges on the surface.
Such methods are typically also inadequate for dealing with noisy 3D points caused by repetitive textures present in documents.
We address these issues by developing a robust ridge-aware surface reconstruction method for sparse 3D points. Specifically, we extend the Poisson surface reconstruction
method~\cite{Kazhdan06} by incorporating ridge constraints and by adding robustness to outliers.

\vspace{3pt}
\noindent\textbf{Robust Poisson surface reconstruction.} We denote a set of $N$ sparse 3D points obtained from SfM as \hbox{$\{\hat{x}_n, \hat{y}_n, \hat{z}_n\}, n = 1, 2, \cdots, N$},
where only 3D points triangulated from at least three images are retained.
For our input images, $N$ typically lies between $700$ to $2000$.
For a selected reference image (and viewpoint), we use a depth map parameterization $z(x,y)$ for the document surface.
We aim to estimate depth at the mesh grid vertices $z_{i}(x_{i}, y_{i})$, where $i$ is the mesh grid index, $1\leq i \leq I$.
Our method computes the optimal depth values $\mathbf{z}^* = \left[z_1, \ldots, z_I \right]^\top$ by
solving the following optimization problem.
\vspace{-4pt}
\begin{equation}
	\mathbf{z}^* = \argmin_{\mathbf{z}} E_{d}(\mathbf{z}) + \lambda E_{s}(\mathbf{z}).
	\label{Eq:Lasso1}
	\vspace{-6pt}
\end{equation}
Here, $E_{d}$ and $E_{s}$ are the data and smoothness terms respectively and $\lambda$ is a parameter to balance the two terms.
The original Poisson surface reconstruction method uses the squared $\ell_2$-norm for both terms. Instead, we propose using
the $\ell_1$-norm for
the data term $E_d$ to deal with outliers.
\vspace{-4pt}
\begin{equation}
	E_{d}(\mathbf{z}) = \sum_{n} \| \hat{z}_n - z_i \|_1.
	\label{Eq:Lasso2}
	\vspace{-6pt}
\end{equation}
This encourages $z_i$ to be consistent with the observed depth $\hat{z}_n$ without requiring explicit knowledge about which observations are outliers. We rewrite \eref{Eq:Lasso2} in vector form.
\vspace{-4pt}
\begin{equation}
	E_{d}(\mathbf{z}) = \|  \hat{\mathbf{z}} - {\cal P}_\Omega \mathbf{z}  \|_1,
	\label{Eq:Lasso2_mat}
	\vspace{-4pt}
\end{equation}
where ${\cal P}_\Omega$ is a permutation matrix that selects and aligns observed entries $\Omega$ by ensuring correspondence between $\hat{z}_n$ and $z_i$.
The smoothness term $E_{s}$ is defined using the squared Frobenius norm of the gradient of depth vector $\mathbf{z}$ along $x$ and $y$ in camera coordinates.
\vspace{-4pt}
\begin{eqnarray}
	E_{s}(\mathbf{z}) = \| \nabla^2 \mathbf{z} \|_F^2 = \left\| \left[ \frac{\partial^2 \mathbf{z}}{\partial x^2},~\frac{\partial^2 \mathbf{z}}{\partial y^2}\right]\right\|_F^2.
	\label{Eq:Lasso3}
	\vspace{-4pt}
\end{eqnarray}

\begin{figure}[t] 
	\centering
	\includegraphics[width=0.9\linewidth]{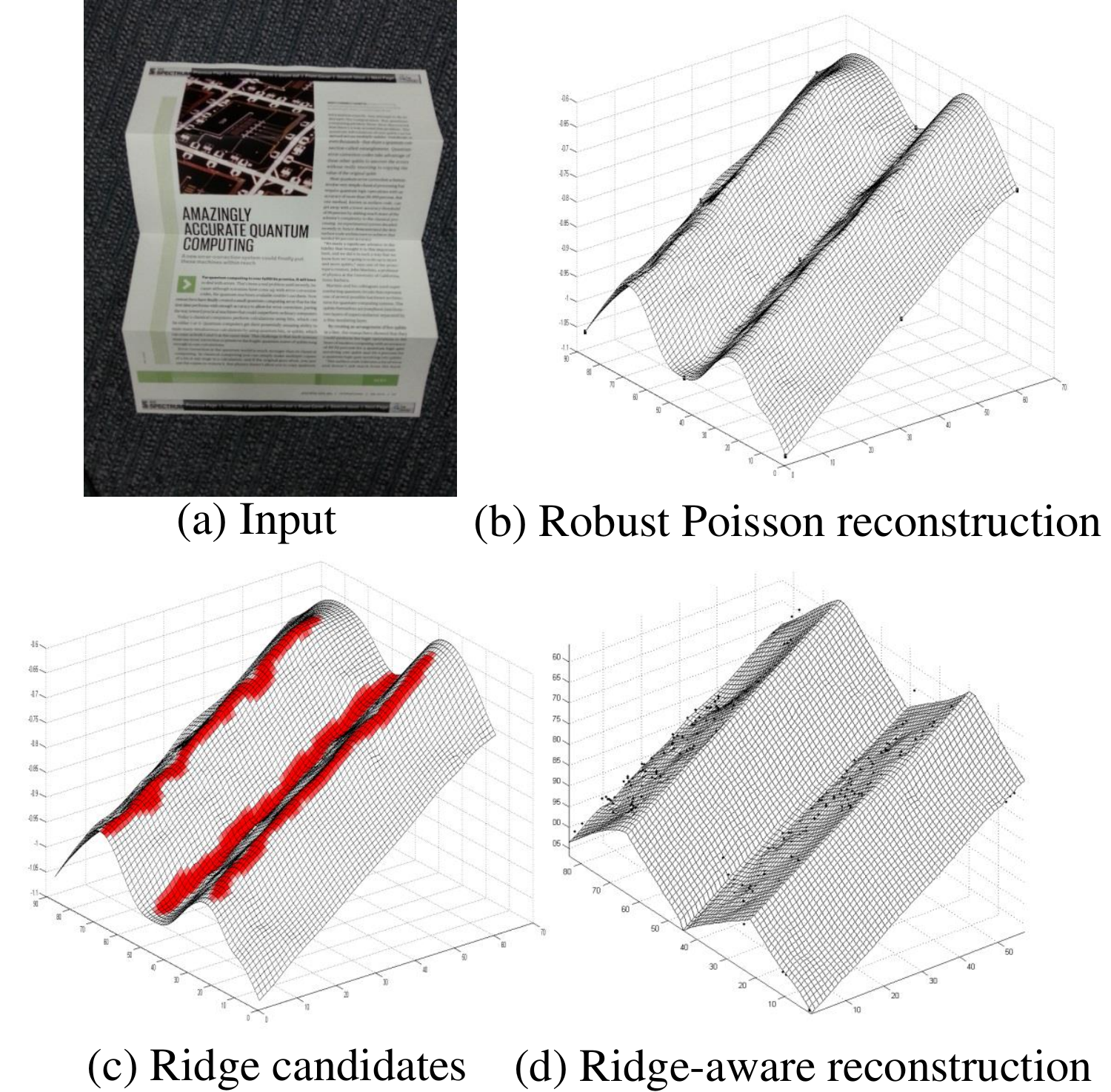}
	\vspace{-6pt}
	\caption{
		\small{
			Example of ridge-aware 3D surface reconstruction.
		}
	}
	\label{Fig:Reconstruction}
	\vspace{-12pt}
\end{figure}

By preparing a sparse derivative matrix $\mathbf{D}$ that replaces the Laplace operator $\nabla^2$ in a linear form,
\vspace{-4pt}
\begin{eqnarray}
	\mathbf{D} =
	\footnotesize{
		\left[
		\begin{split}
			d_{i, j} &=
			\begin{cases}
				2 ~~ & \mathrm{if}~~i = j\\
				-1 ~~ & \mathrm{if}~~z_{j}~\mathrm{is~left/right~to}~z_{i}\\
				0 ~~ & \mathrm{otherwise}
			\end{cases}
			\\
			d_{i+I, j} &=
			\begin{cases}
				2 ~~ & \mathrm{if}~~i = j\\
				-1 ~~ &z_{j}~\mathrm{is~above/below}~z_{i}\\
				0 ~~ & \mathrm{otherwise}
			\end{cases}
		\end{split}
		\right]_{2I \times I}
	},
	\vspace{-4pt}
\end{eqnarray}
we have a special form of the Lasso problem~\cite{Tibshirani96}.
\vspace{-4pt}
\begin{eqnarray}
	\mathbf{z}^* = \argmin_{\mathbf{z}} \| \hat{\mathbf{z}} - {\cal P}_\Omega\mathbf{z} \|_1 + \lambda \| \mathbf{D} \mathbf{z}\|_2^2.
	\label{eq:lasso_form}
	\vspace{-4pt}
\end{eqnarray}
While this problem (\eref{eq:lasso_form}) does not have a closed form solution, we use a variant of iteratively reweighted least squares (IRLS)~\cite{Candes08} for deriving the solution.
By rewriting the data terms in \eref{eq:lasso_form} as a weighted $\ell_2$ norm using a diagonal matrix $\mathbf{W}$ with positive values on the diagonal, we have
\vspace{-4pt}
\begin{eqnarray}
	\!\!\!\mathbf{z}^* \!\!= \!\!\argmin_{\mathbf{z}} \left( \hat{\mathbf{z}} - {\cal P}_\Omega \mathbf{z}  \right)^\top \mathbf{W}^\top \mathbf{W} \left( \hat{\mathbf{z}} - {\cal P}_\Omega \mathbf{z} \right) +  \lambda \mathbf{z}^\top \mathbf{D}^\top \mathbf{D} \mathbf{z}.
	\label{eq:reweight01}
	\vspace{-4pt}
\end{eqnarray}
In contrast to \Eref{eq:lasso_form}, the data term now uses $\ell_2$ norm instead of $\ell_1$ norm.
We solve this problem (\Eref{eq:reweight01}) using alternation as described next.

\vspace{2mm}
\noindent{\emph{Step 1: Update $\mathbf{z}$}}

\noindent \eref{eq:reweight01} can be rewritten as $\mathbf{z}^* = \argmin_{z} \parallel \mathbf{A} \mathbf{z} - \mathbf{b} \parallel _{2}^2$,
where
$
\mathbf{A} =
\left[
\begin{array}{ccc}
\mathbf{W} {\cal P}_{\Omega}\\
\sqrt{\lambda} \mathbf{D}
\end{array}
\right]
$
and
$
\mathbf{b} =
\left[
\begin{array}{ccc}
\mathbf{W} \mathbf{\hat{z}}\\
\mathbf{0}_{2I \times 1}
\end{array}
\right]
$
and $\mathbf{0}_{2I \times 1}$ is a zero vector of length $2I$.
This is a squared $\ell_2$ sparse linear system. It has the closed form solution
\vspace{-4pt}
\begin{eqnarray}
	\mathbf{z}^* = [\mathbf{A}^\top \mathbf{A} + \alpha \mathbf{I}]^{-1} \mathbf{A}^\top \mathbf{b},
	\label{eq:reweight02}
	\vspace{-4pt}
\end{eqnarray}
where $\mathbf{I}$ is the identity matrix, $\alpha$ is a regularization parameter (we use \hbox{$\alpha = 1.0\text{e-}8$}).

\vspace{2mm}
\noindent{\emph{Step 2: Update $\mathbf{W}$}}

\noindent We initialize $\mathbf {W}$ to the identity matrix.
During each iteration, each diagonal element $w_i$ of $\mathbf {W}$ is updated given the residual
\hbox{$\mathbf{r} = \mathbf{W} {\cal P}_{\Omega} \mathbf{z}^* - \mathbf{W} \mathbf{b}$}, as follows.
\vspace{-4pt}
\begin{eqnarray}
	w_i = \frac{1}{|r_i| + \epsilon},
	\label{eq:reweight04}
	\vspace{-4pt}
\end{eqnarray}
Here, $r_i$ is the $i$-th element of $\mathbf{r}$ and $\epsilon$ is a small positive value (we use \hbox{$\epsilon = 1.0\text{e-}8$}).
These steps are repeated until convergence; namely, until the estimate at $t$-th iteration $\mathbf{z}^{*(t)}$ becomes similar to the previous estimate $\mathbf{z}^{*(t-1)}$, \ie, $\|\mathbf{z}^{*(t)} - \mathbf{z}^{*(t-1)}\|_2 < 1.0\text{e-}8$.
\Fref{Fig:Reconstruction}.c shows an example of the reconstructed mesh.

\vspace{3mm}
\noindent\textbf{Ridge-aware reconstruction.} Developable surfaces are ruled~\cite{Portnoy75}, \ie, contain straight lines on the surface as shown in~\fref{Fig:Ruler}.
Our method exploits this geometric property as described in this section.
Unlike existing parameterization-based methods which only handle smooth rulers, \cite{Liang08, Tian11, Meng14, Perriollat13}, extracting arbitrary creases and ridges is more difficult when only sparse 3D points are available. We propose a sequential approach by first detecting ridges on the mesh ${\mathbf{z}^*}$ that was obtained using our robust Poisson reconstruction method. After selecting the ridge candidates, we instantiate additional linear ridge constraints and incorporate them into the linear system that was solved earlier. This sequential approach is quite general and avoids overfitting. It also avoids spurious ridge candidates arising due to noise.

For each point $z(x, y)$ on the mesh $\mathbf{z}^*$, we compute the Hessian $\mathbf{K}$ as follows.
\vspace{-4pt}
\begin{equation}
	\mathbf{K}(z) =
	\left[
	\begin{array}{ccc}
		\frac{\partial^2 z}{\partial x^2} & \frac{\partial^2 z}{\partial x \partial y}\\
		\frac{\partial^2 z}{\partial x \partial y} & \frac{\partial^2 z}{\partial y^2}
	\end{array}
	\right].
	\label{Eq:Ridge1}
	\vspace{-4pt}
\end{equation}
Based on the following Eigen decomposition of $\mathbf{K}(z)$,
\vspace{-4pt}
\begin{equation}
	\mathbf{K}(z) =
	\left[ \mathbf{p}_1, \mathbf{p}_2 \right]
	\left[
	\begin{array}{ccc}
		\kappa_1 & 0 \\
		0 & \kappa_2
	\end{array}
	\right]
	\left[ \mathbf{p}_1, \mathbf{p}_2 \right]^\top,
	\label{Eq:Ridge2}
	\vspace{-4pt}
\end{equation}
we obtain principal curvatures $\kappa_1$ and $\kappa_2$ (\hbox{$|\kappa_1| \leq |\kappa_2|$}) and the corresponding eigenvectors $\mathbf{p}_1$ and $\mathbf{p}_2$.

The value of $\kappa_1$ is equal to zero at all points on a developable surface. Thus,
at any point $z_i$, a straight line along direction $\mathbf{p}_1$ must lie on the surface.
As discussed earlier and shown in \fref{Fig:Ruler}, the curvature along the ridge is zero while the curvature orthogonal to the ridge reaches a local extremum.
We use this observation to select ridge candidates using the value of $|\kappa_2|$.
Mesh points $z_{i}(x_i, y_i)$ with $|\kappa_2(i)|$ greater than the threshold $\kappa_{th}$ are selected as ridge candidates.
(see \Fref{Fig:Reconstruction}d for an example).
The associated smoothness constraints in \eref{Eq:Lasso3} are adjusted as follows.
\vspace{-4pt}
\begin{eqnarray}
	\begin{split}
		\tilde{d}_{i, j} &= \varphi ( \langle \mathbf{p}_1, \mathbf{e}_1 \rangle ) d_{i, j}
		\\
		\tilde{d}_{i+I, j} &= \varphi ( \langle \mathbf{p}_1, \mathbf{e}_2 \rangle ) d_{i + 1, j},
	\end{split}
	\label{Eq:Ridge3}
	\vspace{-4pt}
\end{eqnarray}
where $\langle ~, ~ \rangle$ is the inner product and $\mathbf{e}_1 = [1, 0]^\top, \mathbf{e}_2 = [0, 1]^\top$ are orthonormal bases. $\varphi(\cdot)$ is a convex monotonic function defined as
\hbox{$\varphi(x) = \frac{\beta^{x^2}-1}{\beta-1}$},
which places a greater weight \hbox{$\beta \gg 1$} along the ridge and smaller weight orthogonal to it.
We also consider two more directional smoothness constraints similar to those stated in \eref{Eq:Ridge3}, for the two diagonal directions
\hbox{$\mathbf{e}_3 = [\frac{\sqrt{2}}{2}, \frac{\sqrt{2}}{2}]^\top$} and \hbox{$\mathbf{e}_4 = [\frac{\sqrt{2}}{2}, -\frac{\sqrt{2}}{2}]^\top$}.

Finally, we modify $E_{s}(\mathbf{z})$ defined in \eref{Eq:Lasso3} by adding these ridge constraints and solve a new sparse linear system (similar to the earlier one)
to obtain the final reconstruction. \Fref{Fig:Reconstruction}e shows that this method can preserve accurate folds and creases.

\vspace{-6pt}
\subsection{Surface Unwrapping}

Given the 3D surface reconstruction, our next step is to unwrap the surface.
We take a conformal mapping approach to this problem, amongst which,
Least Squares Conformal Mapping (LSCM)~\cite{Levy02,Brown07} is a suitable choice.
However, it is not resilient to the presence of outliers and susceptible to global distortion which can occur due to the absence of long-range constraints.
We address both these issues and extend LSCM by incorporating an appropriate robustifier as well as ridge constraints to reduce global drift.

\begin{figure}[tbhp]
	\centering
	\includegraphics[width=0.9\linewidth]{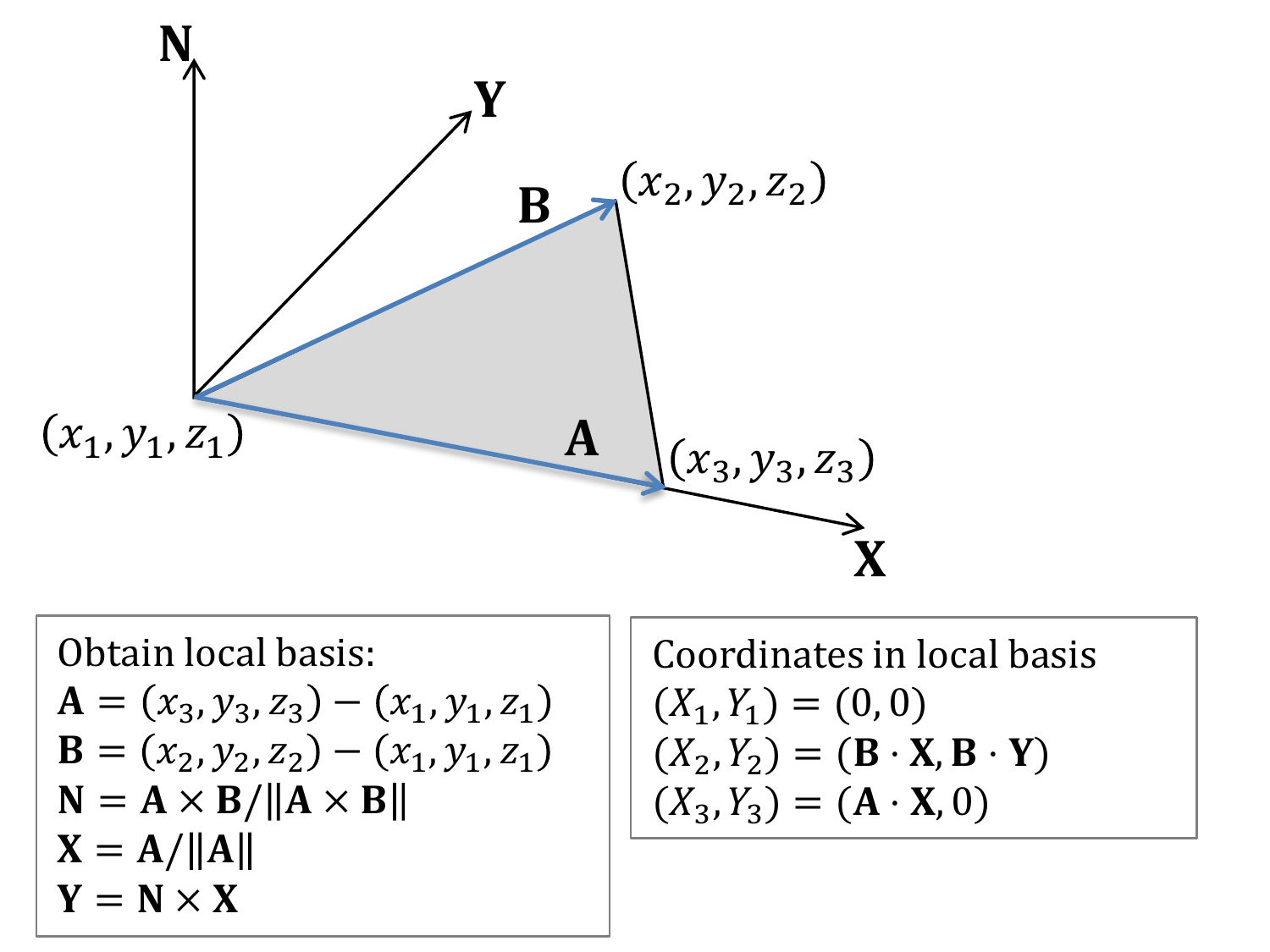}
	\vspace{-6pt}
	\caption{
		\small{
        Vertices of a triangle in a local coordinate basis.
    }
	}
	\label{Fig:Conformal}
	\vspace{-12pt}
\end{figure}
\vspace{1mm}
\noindent\textbf{Conformal Mapping.} For our mesh topology, each 3D point $z_{i}(x_i, y_i), i = 1, \dots, I$, on the grid on $\mathbf{z}$
forms two triangles, one with its upper and left neighbor, the other with its lower and right neighbor on the grid. The triangulated 3D mesh is denoted as $\{ \mathbf{\cal T}, \mathbf{z}\}$.
A conformal map will produce an associated 2D mesh with the same connectivity but with 2D vertex positions such that the angle of all the triangles are best preserved.
We denote the 2D mesh as $\{ \mathbf{\cal T}, \mathbf{u}\}$, where $\mathbf{u} = (u_i, v_i)$.

For a particular 3D triangle $t$ with vertices at \hbox{$(x_1, y_1, z_1)$}, \hbox{$(x_2, y_2, z_2)$}, and \hbox{$(x_3, y_3, z_3)$}, we seek its associated 2D vertex positions (\hbox{$(u_1, v_1)$}, \hbox{$(u_2, v_2)$} and \hbox{$(u_3, v_3)$}) under the conformal map. Using a local 2D coordinate basis for triangle $t$, the conformality constraint is captured by the following linear equations.
\vspace{-9pt}
\begin{eqnarray}
	\frac{1}{S}
	\tiny{
		\left[
		\begin{array}{cccccc}
			\Delta X_1 & \Delta X_2 & \Delta X_3 & -\Delta Y_1 & -\Delta Y_2 & -\Delta Y_3\\
			\Delta Y_1 & \Delta Y_2 & \Delta Y_3 & \Delta X_1 & \Delta X_2 & \Delta X_3
		\end{array}
		\right]
	}
	\mathbf{u}_{t}
	= \mathbf{0},
	\label{Eq:Conformal01}
	\vspace{-4pt}
\end{eqnarray}
Here, \hbox{$\mathbf{u}_t = [u_1, u_2, u_3, v_1, v_2, v_3]^\top$}, $S$ is the area of $t$,
\hbox{$\Delta X_1 = (X_3 - X_2)$}, \hbox{$\Delta X_2 = (X_1 - X_3)$} and \hbox{$\Delta X_3 = (X_2 - X_1)$} ($\Delta Y$ is similarly defined).
Note that variables \hbox{$(X_1, Y_1)$}, \hbox{$(X_2, Y_2)$}, and \hbox{$(X_3, Y_3)$} were obtained from $t$'s vertex coordinates (see \fref{Fig:Conformal}).
Putting together the constraints for all the triangles, we have the following sparse linear system.
\vspace{-3pt}
\begin{eqnarray}
	\mathbf{C} \mathbf{u} = \mathbf{0}.
	\label{Eq:Conformal02}
	\vspace{-6pt}
\end{eqnarray}
Using indices $i$ and $j$ to index the $I$ vertices and $J$ triangles respectively,
the $2J \times 2I$ matrix $\mathbf{C}$ in \Eref{Eq:Conformal02} has the following non-zero entries.
\vspace{-2pt}
\begin{eqnarray}
	\begin{array}{cc}
		c_{j, i}= \frac{\Delta X}{S_{j}},
		&
		c_{j, i + I}= -\frac{\Delta Y}{S_{j}}
		\\
		c_{j+J, i}=\frac{\Delta Y}{S_{j}},
		&
		c_{j + J, i + I}=\frac{\Delta X}{S_{j}}
	\end{array}
	\label{Eq:Conformal03}
    \vspace{-4pt}
\end{eqnarray}

\begin{figure}[tbp]
	\centering
	\includegraphics[width=\linewidth]{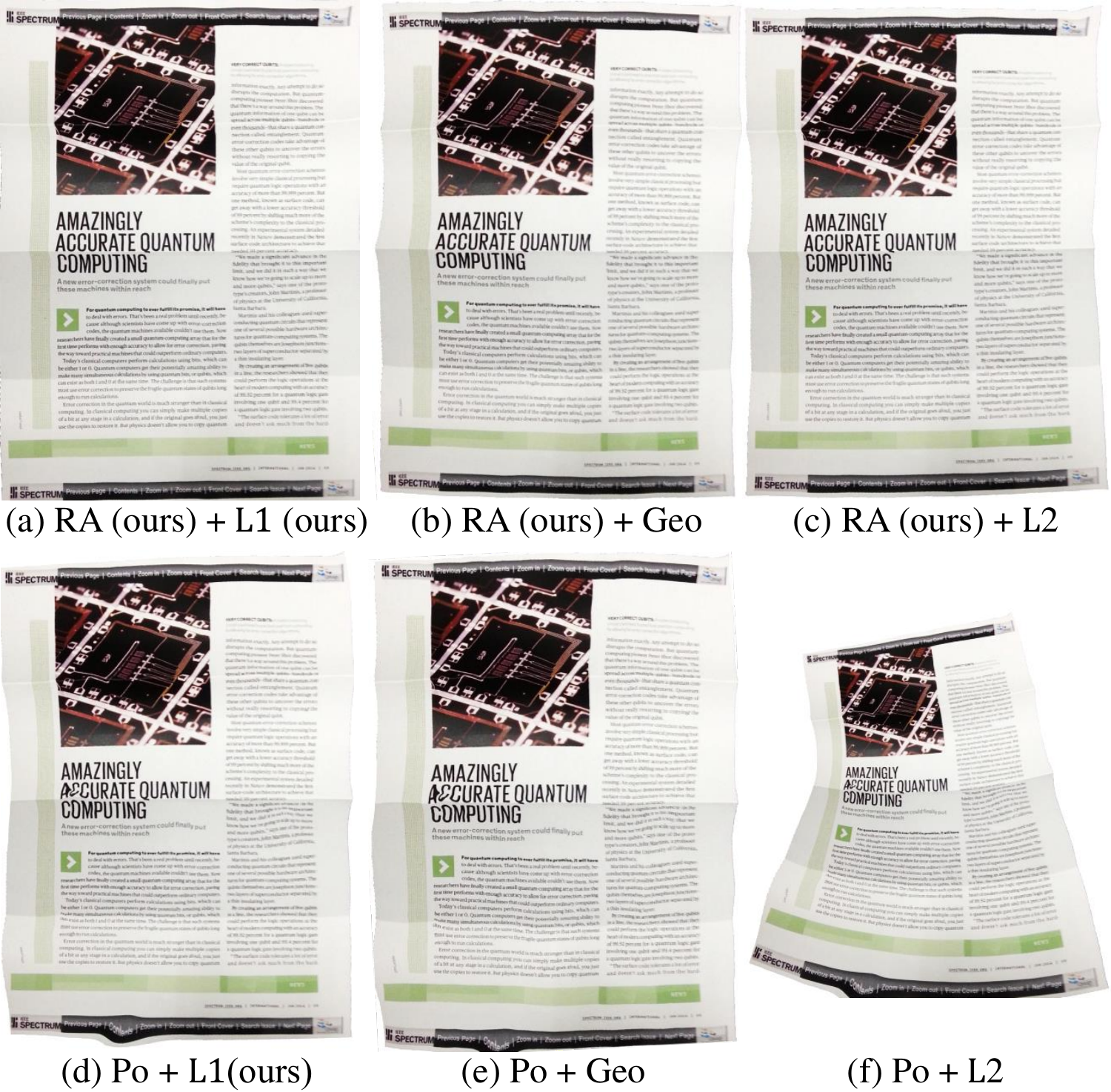}
	\vspace{-12pt}
	\caption{
		\small{
			Rectification results from combination of methods.
			Acronyms RA and Po denote our ridge-aware method and Poisson reconstruction respectively. L1 denotes our $\ell_1$ conformal mapping method with non-local constraints; L2 indicates LSCM~\cite{Brown07} and Geo indicates geodesic unwrapping~\cite{Zhang08}. Po + L2 produced gross failures and is thus not compared.
		}
	}
	\label{Fig:RVisual}
	\vspace{-18pt}
\end{figure}

\vspace{0pt}
\noindent\textbf{Ridge constraints.} Notice that the original conformal mapping has only local constraints, which will result in global distortion, \fref{Fig:RVisual}.f. To reduce global distortions during unwrapping, we add ridge and boundary constraints to constrain the solution further.

We take into consideration two facts. First, the ridge lines remain straight after flattening but should essentially become invisible on the flattened surface. Second, the conformal mapping constraint \eref{Eq:Conformal01} applies to beyond triangles. In particular, it is true for three collinear points.
Therefore, we propose using the collinearity property to derive non-local constraints during flattening.
and add it to our conformal map estimation problem.
Referring to \fref{Fig:Conformal}, and imagine the collinear case, that is when point $(x_2, y_2, z_2)$ is also lying on the $\mathbf{X}$ axis; in such case, $Y_2=Y_1=Y_3=0$. In addition, the area of the triangle $T$ is zero. 
Hence, the ridge constraints can be written in a form similar to \eref{Eq:Conformal01}.
\vspace{-4pt}
\begin{eqnarray}
	\footnotesize{
		\left[
		\begin{array}{cccccc}
			\Delta X_1 & \Delta X_2 & \Delta X_3 & 0 & 0 & 0 \\
			0 & 0 & 0 & \Delta X_1 & \Delta X_2 & \Delta X_3
		\end{array}
		\right]
	}
	\mathbf{u}_{\cal R}
	= \mathbf{0}.
	\label{Eq:Conformal04}
	\vspace{-4pt}
\end{eqnarray}
where $\mathbf{u}_{\cal R} = [u_1, u_2, u_3, v_1, v_2, v_3]^\top$ are the targeted 2D coordinates similarly defined as $\bm{u}_t$.
We select ridge candidates in the same way as we did earlier during reconstruction. However, this step is now more accurate because the surface is well reconstructed.
For each ridge candidate (vertex), we find two farthest ridge candidates along the ridge line in opposite directions and instantiate the above mentioned constraint for the three points.
We assume that the boundary of the flattened 2D document image has straight line segments (they need not be straight lines on the 3D surface).
These boundary constraints can be expressed in a form similar to \eref{Eq:Conformal04}.
We incorporate all ridge and boundary constraints into a system of linear equations.
\vspace{-2pt}
\begin{eqnarray}
	\mathbf{R} \mathbf{u} = \mathbf{0}.
	\label{Eq:Conformal05}
	\vspace{-8pt}
\end{eqnarray}


\vspace{-6pt}
\noindent\textbf{Robust conformal mapping.} We propose using an $\ell_1$ norm instead of the standard squared $\ell_2$ norm
to make conformal mapping robust to outliers. Putting together Eqs. (\ref{Eq:Conformal02}) and (\ref{Eq:Conformal05}) in the $\ell_1$ sense, we have
\vspace{-0pt}
\begin{eqnarray}
	\mathbf{u}^* = \argmin_{\mathbf{u}} \parallel \mathbf{C} \mathbf{u} \parallel_1 + \gamma  \parallel \mathbf{R} \mathbf{u} \parallel_1,
	\label{Eq:Conformal06}
	\vspace{-18pt}
\end{eqnarray}
where $\gamma$ balances the ridge and boundary constraints.
To avoid the trivial solution $\mathbf{u} = 0$, we fix two points of $\mathbf{u}$ to $(u_i, v_i) = (0, 0)$ and $(u_j, v_j) = (0, 1)$. \Eref{Eq:Conformal06} is then rewritten as
\vspace{-4pt}
\begin{eqnarray}
	\mathbf{u}^* = \argmin_{\mathbf{u}} \parallel \mathbf{C} \mathbf{u} \parallel_1
	+ \gamma  \parallel \mathbf{R} \mathbf{u} \parallel_1
	+ \theta  \parallel E_{\rm{fix}} \parallel_2^2,
	\label{Eq:Conformal07}
	\vspace{-9pt}
\end{eqnarray}
where $E_{\rm{fix}}$ is the energy function for the two fixed points.
We solve the objective function using the iterative reweighted least squares method~\cite{Candes08}.
\Fref{Fig:RVisual} shows a result from the conventional LSCM ($\ell_2$ method) and our proposed $\ell_1$ method.

\vspace{-4pt}
\subsection{Implementation details}
\vspace{-2pt}

\noindent\textbf{Sparse 3D reconstruction.} We recover the initial sparse 3D point cloud using SfM.
While any existing SfM method is applicable, we use the popular incremental SfM technique~\etal~\cite{Hartley04} in our system.
We typically capture five to ten still images for each document from different viewpoints. Capturing these images or
equivalently a set of burst photos or a short video clip only takes a few seconds.
\Fref{Fig:Pipeline}a shows an input example and the corresponding reconstruction.

\vspace{1pt}
\noindent\textbf{Image warping.}
After recovering the flattened mesh grid $\mathbf{u} = \{u_i, v_i\}$, we unwrap the input image with the maximum document area in pixels. To obtain correspondence between the input image and $\{u_i, v_i\}$, we project the 3D mesh points $\{z_i(x_i, y_i)\}$ into the image to obtain image coordinates $\{\tilde{x_i}, \tilde{y_i}\}$ using the camera pose estimated using SfM.
We then warp the image according to the correspondence between $\{\tilde{x_i}, \tilde{y_i}\}$ and $\{u_i, v_i\}$ with bilinear interpolation.

\begin{figure*}[tbhp]
	\centering
	\includegraphics[width=\linewidth]{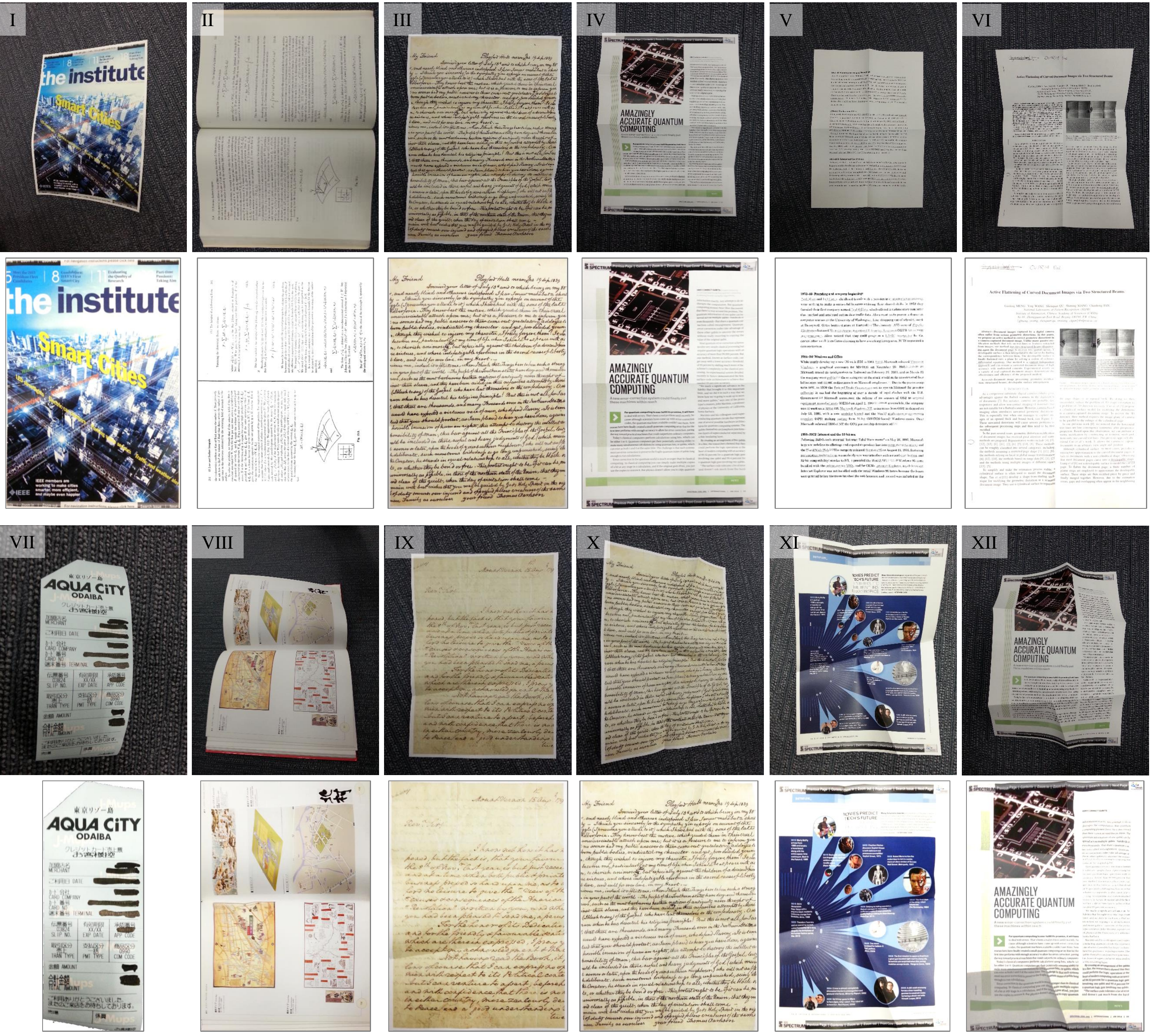}
	\vspace{-18pt}
	\caption{
		\small{
		[\textsc{Our results}] Original images are shown in rows 1 and 3 and our rectification results are shown in rows 2 and 4.
	}
	}
	\vspace{-15pt}
	\label{Fig:Visual}
\end{figure*}

\section{Experiments}

We perform qualitative and quantitative evaluation on a wide variety of input documents. The first set of experiments show that our method can handle different paper types, document content and various types of folds and creases. Next, we report a quantitative evaluation based on known ground truth using local and global metrics where we demonstrate the superior performance and advantages of our method over existing methods~\cite{Brown07, Zhang08,Perriollat13}. In all the experiments, we set parameters as follows: $\lambda=1\text{e-}5, \beta=40,$ $\gamma=1e3$, $\theta=1e2$ and $\kappa_{th}=0.006$.
Our method is insensitive to these parameters. Varying $\lambda, \gamma, \theta$ by factors of 0.1 -- 1.0 or varying $\beta$ or $\kappa_{th}$ by 50\% from these settings did not change the result significantly.

\vspace{-2pt}
\subsection{Test Data}

The first six out of the 12 test sequences (I -- VI) contain documents with no fold lines, one fold line, two to three parallel fold lines, and two to three crossing fold lines respectively. The other six sequences (VII -- XII) contain documents with an increasing number of fold lines. Irregular fold lines were intentionally added to make the rectification more challenging.
All documents were either placed on a planar or curved background surface. Sequence VII contains a shopping receipt on a paper roll whereas II and VIII contain pages from a book. Sequences III, IV, IX and X contain letters folded within envelopes. Sequence V, VI, XI and XII contain examples of documents folded inside a purse or notebook.
The input images as well as the results from our method are shown in \fref{Fig:Visual}.
Our method does not rely on the content, formatting, layout or color of the document. Thus, it is generally applicable as long as a sufficient number of sparse keypoints in the input images are available for SfM.

\begin{figure}[tbhp]
	\centering
	\includegraphics[width=\linewidth]{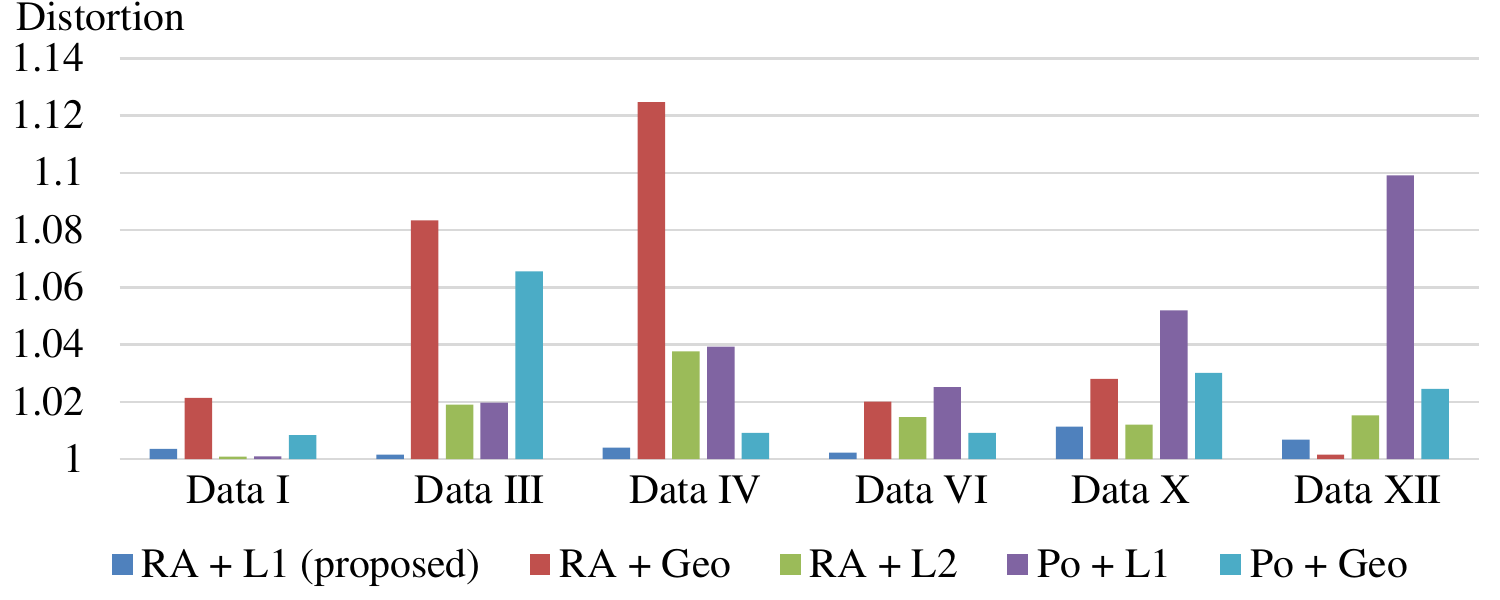}
	\vspace{6pt}
	\footnotesize{Global distortion evaluation metric}
	\includegraphics[width=\linewidth]{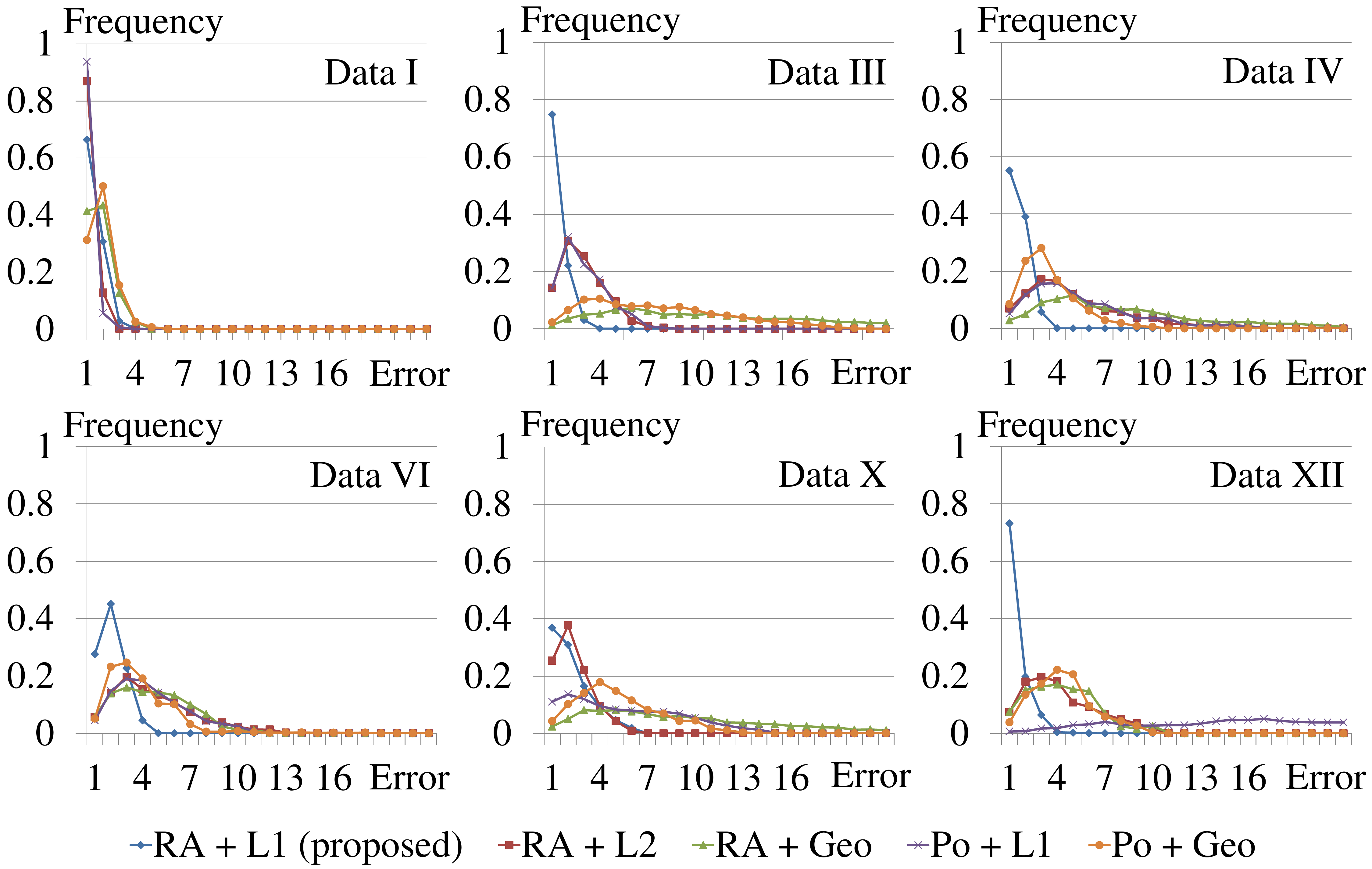}
	\footnotesize{Local distortion evaluation metric}
	\vspace{-6pt}
	\caption{
		\small{
		Distortion metrics for datasets shown in \fref{Fig:Visual}. 			 Abbreviations are consistent with \fref{Fig:RVisual} and the text.
		}
	}
	\label{Fig:RGlobal}
	\vspace{-18pt}
\end{figure}

\subsection{Quantitative Evaluation Metrics}
We quantitatively evaluate the global and local distortion between the ground truth digital image and our rectified result using local and global metrics.
The digital version of six out of the 12 test documents were available. We treat those images as ground truth and resize them by setting their
height to 1000 pixels.

\vspace{2mm}
\noindent\textbf{Global distortion metric.}
We first register the rectified image to the ground truth by estimating a global affine transform $T$ estimated using SIFT keypoint correspondences in these images~\cite{Lowe04}.
\begin{eqnarray}
	\mathbf{T} = \left[
	\begin{small}
	\begin{array}{ccc}
		a_1 & a_2 & t_1 \\
		a_3 & a_4 & t_2 \\
		0 & 0 & s
	\end{array}
	\end{small}
	\right],
	\label{Eq:Trans01}
	\vspace{-8pt}
\end{eqnarray}
This is achieved by minimizing the squared error.
\vspace{-4pt}
\begin{eqnarray}
	\mathbf{T}^* = \argmin_{\mathbf{T}} \parallel \mathbf{T}\mathbf{p} - \mathbf{\hat{p}} \parallel_2^2.
	\label{Eq:Trans02}
	\vspace{-8pt}
\end{eqnarray}
where, $\mathbf{p}$ and $\mathbf{\hat{p}}$ denote corresponding 2D keypoint positions using homogenous coordinates.
We compute the global distortion metric $\mathcal{G}$ as follows.
\vspace{-3pt}
\begin{eqnarray}
	\begin{split}
		G &= (a_1 a_4 - a_2 a_3) / s^2\\
		\mathcal{G} &= \max{(G, 1 / G}).
	\end{split}
	\label{Eq:Trans03}
	\vspace{-6pt}
\end{eqnarray}
A perfect result has $\mathcal{G} = 1$; and larger values indicate more distortion
(see comparative results in \fref{Fig:RGlobal}).

\vspace{1pt}
\noindent\textbf{Local distortion metric.} We compute the local metric by performing dense image registration using SIFT-flow \cite{Liu08} between the rectified image and the ground truth image. The frequency distribution of local displacements are shown in lower figure in \fref{Fig:RGlobal} and compared with existing methods. We found dense registration to be more useful
for an unbiased evaluation than sparse SIFT keypoint-based registration because sparse methods are more likely to ignore many matches if the result contains large deformations.

\subsection{Comparison with existing methods}

We first compare with three methods~\cite{Perriollat13,Brown07,Zhang08} on various real  images  and  then  use  synthetically generated  data  to further compare with  the methods  designed  for  dense  3D  point data ~\cite{Brown07,Zhang08}.

\begin{figure}[tb]
	\centering
	\includegraphics[width=\linewidth]{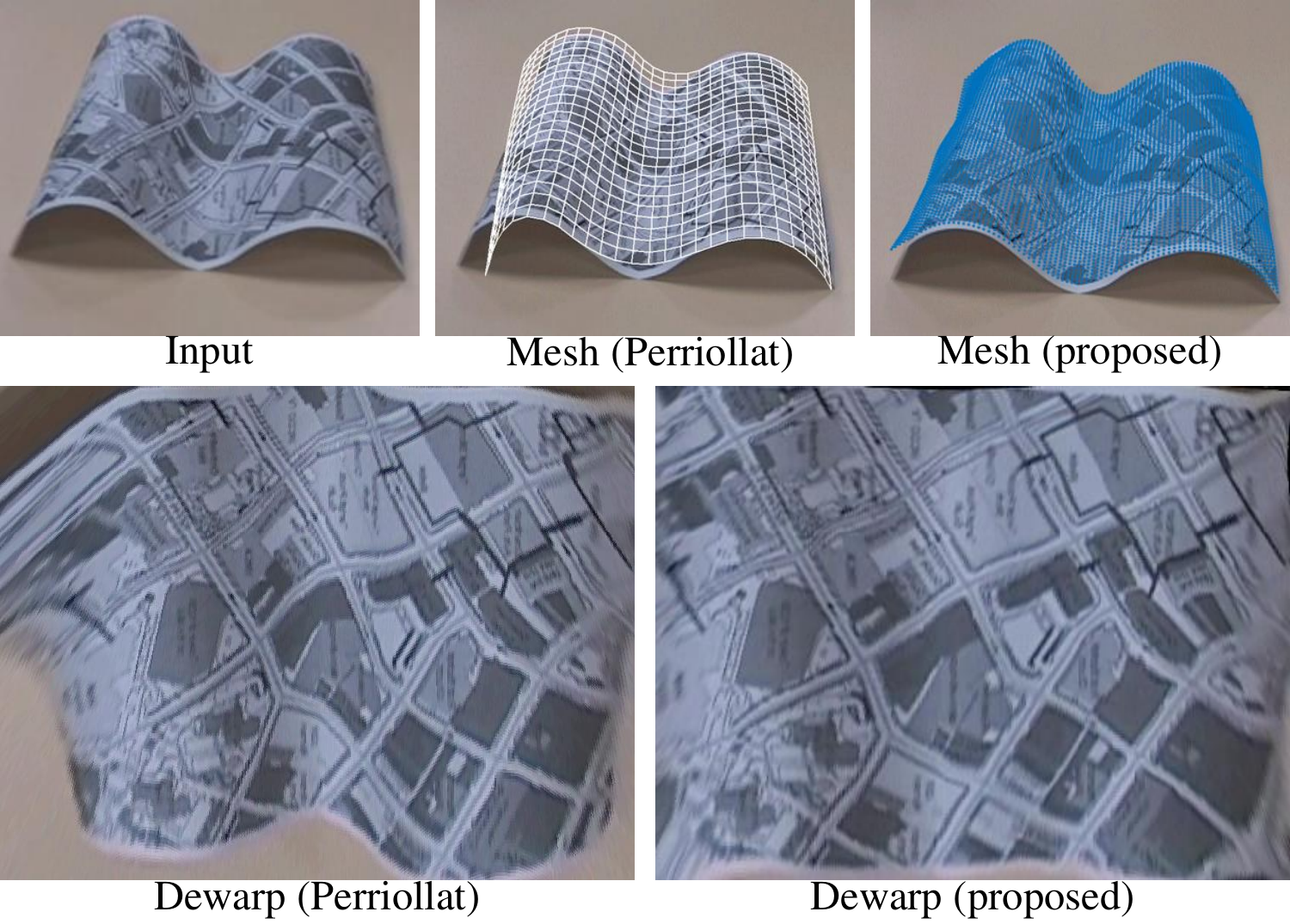}
	\vspace{-24pt}
	\caption{
		\small{
		Visual comparison with cylindrical like surface based method which uses SfM as input \cite{Perriollat13}.
		}
	}
	\label{Fig:Perrio}
	\vspace{-12pt}
\end{figure}

\vspace{2mm}
\noindent \textbf{Perriollat \etal~\cite{Perriollat13}.}
Their method explicitly parameterizes smooth rulers but cannot handle our document images with creases and folds.
Our method works fine on their dataset and produces a more accurate result than the one obtained by running their code
\footnote{Their result shown here was generated by the original code provided by the authors. These result do not agree with the results in their paper. This is probably due to a difference in initialization.}
 (see \fref{Fig:Perrio}). Although our result has minor artifacts due to self-occlusion and fore-shortening, the flattening result is quite accurate.

\vspace{2mm}
\noindent \textbf{Brown \etal~\cite{Brown07}, Zhang \etal~\cite{Zhang08}.}
We compare to both methods using our sequences where ground truth is available (\fref{Fig:Visual}).
Since they require 3D range data, we use our reconstructed surface as their input and compare the surface flattening quality. We also compare our ridge-aware reconstruction to the standard Poisson reconstruction method. As shown in \fref{Fig:RGlobal}, the global and local distortion metrics introduced earlier are used in the evaluation. Our method has higher accuracy in terms of both metrics.
Results from various methods have been compared in \fref{Fig:RVisual}. 

\begin{figure*}[tbp]
	\centering
	\includegraphics[width=\linewidth]{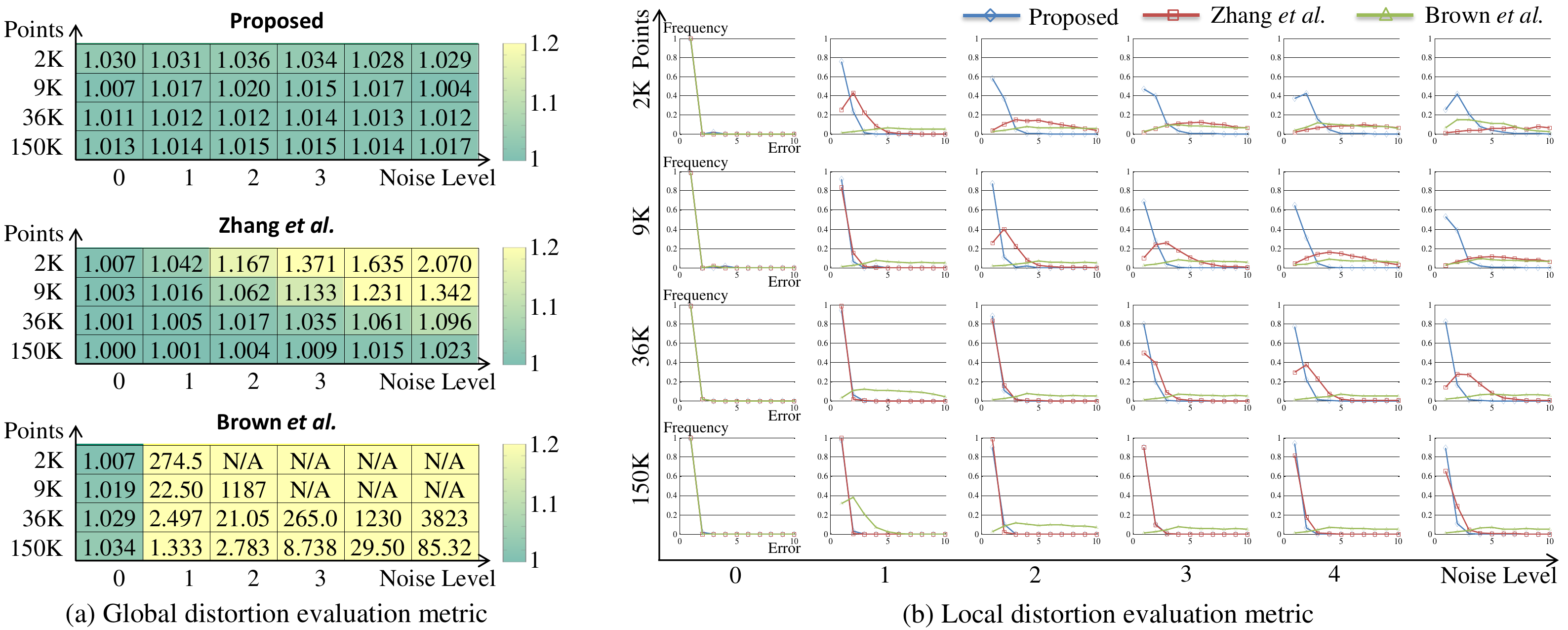}
	\vspace{-18pt}
	\caption{
		\small{
			Comparison of the global distortion metric between our method (top) and Zhang~\etal~\cite{Zhang08} and Brown~\etal\cite{Brown07} with varying point density and noise. Here lower values indicate higher accuracy. (b) Frequency distribution of local distortion metrics for the associated experiments. Our method is more accurate when input point are sparser or have more noise.
		}
	}
	\vspace{-18pt}
	\label{Fig:SGlobal}
\end{figure*}

\vspace{2mm}
\noindent \textbf{Evaluation on synthetic data.}
We compared our method with~\cite{Brown07,Zhang08} on synthetically generated dense 3D points
because these methods require dense 3D points. We vary the point cloud size from 2K to 300K (common in 3D range data) and inject varying levels of Gaussian noise.
The results from the three methods are compared in \Fref{Fig:SGlobal}. These experiments show that with low noise and high point density,
all three methods are comparable in accuracy. However, when the points are sparser or when the noise level is higher, our method is more
accurate than prior methods~\cite{Brown07,Zhang08}.

\vspace{-0pt}
\section{Conclusion and Future Work}
\vspace{-0pt}
In this paper, we propose a method for automatically rectifying curved or folded paper sheets from a small number of images captured from different viewpoints.
We use SfM to obtain sparse 3D points from images and propose ridge-aware surface reconstruction method which utilizes the geometric property of developable surface for accurate and dense 3D reconstruction of paper sheets.
We also robustify the algorithms using $\ell_1$ optimization techniques.
After recovering surface geometry, we unwrap the surface by adopting conformal mapping with both local and non-local constraints in a robust estimation framework.
In the future we will address the correction of photometric inconsistencies in the document image caused by shading under scene illumination.

\end{document}